\documentclass{isprs} % isprs class modified 23-04-2019 (Dennis Wittich)
\usepackage[utf8]{inputenc}
\usepackage{tikz}
\usepackage{subcaption}
\usepackage{graphicx}
\usepackage{bbm}
\usepackage{amsmath}
\usepackage{amsfonts}
\usepackage{amssymb}
\usepackage{setspace}
\usepackage{geometry} % added 27-02-2014 Markus Englich
\usepackage{epstopdf}
\usepackage[labelsep=period]{caption}  % added 14-04-2016 Markus Englich - Recommendation by Sebastian Brocks
\usepackage[british]{babel}

\DeclareMathOperator*{\argmin}{arg\,min}
\title{Automatic clustering of Celtic coins based on 3D point cloud pattern
	analysis}

\author{
	Sofiane Horache\textsuperscript{1}, François Goulette\textsuperscript{1}, Jean-Emmanuel Deschaud \textsuperscript{1}, Thierry Lejars\textsuperscript{2}, Katherine Gruel\textsuperscript{2}}
\address{
	\textsuperscript{1 }Centre de Robotique,  CAOR, Mines Paristech, PSL University (name.surname)@mines-paristech.fr\\
	\textsuperscript{2 }- AOROC, ENS Ulm, PSL University((name.surname)@ens.fr)\\
	
}

\commission{II}{}
\workinggroup{II/8}
\icwg{}   %This field is optional.

\abstract{
	The recognition and clustering of coins which have been struck by the
	same die is of interest for archeological studies. Nowadays, this work
	can only be performed by experts and is very tedious.
	In this paper, we propose a method to automatically cluster dies, based
	on 3D scans of coins. It is based on three steps: registration,
	comparison and graph-based clustering.
	Experimental results on 90 coins coming from a Celtic treasury from
	the II-Ith century BC show a clustering quality
	equivalent to expert's work.
}


\begin{document}

\maketitle

%\saythanks % added 28-02-2014 Markus Englich

\section{INTRODUCTION}\label{INTRODUCTION}

\begin{figure}[ht]
	\includegraphics[width=0.5\textwidth]{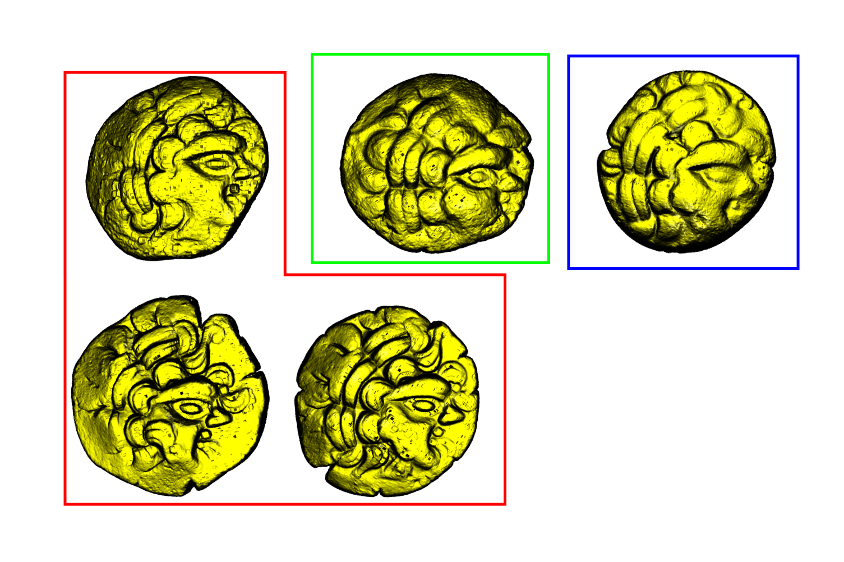}
	\caption{Example of 3D point cloud model of coins. The 3 coins framed in red have the same die. The two other coins have been struck each by a different die. You can see the similarity of the pattern for all coins.}
	\label{fig:exp_man}
\end{figure}

The Celts are well-known for the high quality of their metallurgical productions (shields, weapons, jewels, helmets, coins, and so on). In order to better understand their manufacturing methods, archaeologists rely on detailed observations of objects, in which 3D scans are becoming increasingly important. Celtic coins can give very useful information to better understand the socio-economic context. Coins are among the first mass-produced objects with repeatability constraints imposed on composition, shape, size, weight and image. The image is the mark of the issuing power. For archaeologists, detailed analysis of coins are therefore crucial \cite{richard2014celt,gruel1981celt,brousseau2009celt}. One aim is to estimate the number of dies\footnote{A die is a stamping tool containing the image of the pattern. With a high pressure, pattern can be stamped in the coin} used in the process of fabrication to estimate the volume of money supply.
To do so, archaeologists compare the pattern of coins and try to see if they are struck by the same die.
As we can see in Figure \ref{fig:exp_man}, the problem is difficult because the differences between patterns are hardly perceptible for an untrained eye (the coins are about 2 cm wide).
 Moreover, the coins may be damaged, worn or even fractured. Archaeologists can find thousands of coins and it would take too much time to compare every pair of coins. That's why, they need automatic tools to help them to know whether the coins come from the same die or not. To solve this challenging task, we propose a rather simple but effective method to know automatically whether two coins come from the same die and to cluster them.
To recognize dies, we first align the patterns using a registration algorithm. Then, by analyzing the histograms of point to point distance maps, we can tell whether two coins are struck by the same dies or not. With all these comparisons, it is possible then to cluster coins that have the same die.
% TODO style
Registration algorithms are algorithms which align 3D point cloud and are usually used for reconstruction, SLAM (Simultaneous Odometry And Mapping) or even 3D object tracking \cite{vongkulbhisal_discriminative_2017}. But aligning partially scanned scenes or 3D models and aligning patterns are different. Indeed, registration algorithms usually align shapes, whereas in our problem, we want to align patterns in arbitrary shapes. That's why we need to adapt registration algorithms to perform this task.

Our contribution is the following:
\begin{itemize}
	\item We propose a method that performs accurate coin die clustering with small annotated dataset
\end{itemize}
In order to realize this objective, we adapted the registration algorithm for the pattern alignment, and we also define a metric to know whether two patterns are similar. 
Extensive experiments have been performed to show the interest of our method.

\section{RELATED WORK}\label{sec:RELATED WORK}

\subsection{Pattern recognition applied on point cloud} \label{sec:Pattern Recognition}
Researchers like \cite{zambanini_classifying_2014,zambanini_coarse_fine_2013,schlag_ancient_2017,salgado_medieval_2016} usually work with images because it is easier to acquire images and pattern recognition techniques work very well on images. According to \cite{salgado_medieval_2016} and to our knowledge, no work
%TODO quote thesis
 has been published on 3D scans of coins for classification or clustering. Moreover, researchers usually try to get a rough classification of the very different patterns. But in our dataset, coins have the same class of patterns and we want to know if they come from the same die or not (a sample of our dataset can be found in Figure \ref{fig:exp_man}).
 \begin{figure}[ht]
 	\centering
 	\includegraphics[width=0.3\textwidth]{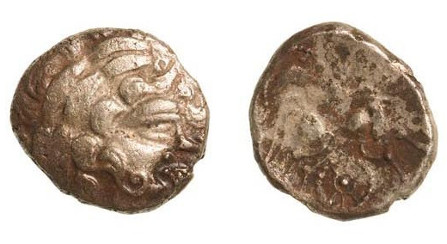}
 	\caption{Example of a RGB image of both side of a coin.}
 	\label{fig:real_coin}
 \end{figure}
 In our problem, it is much harder to use RGB images.  In Figure \ref{fig:real_coin}, we can see rust in the coins and other color artifacts. With 3D scanner, we have access to the geometry of the pattern. Therefore, it seems to us that it is easier to solve the problem with very accurate 3D data instead of images. \cite{salgado_medieval_2016} performs classification of coins with descriptors, using a popular method like SIFT \cite{lowe_distinctive_2004}. With descriptors, they can use Bag of Visual Word for each image and use SVM for the classification.
\cite{schlag_ancient_2017} use Deep Learning to perform classification on image. 
But direct classification is ineffective in our case because we don't know in advance the number of classes and, we don't have so many labeled data. To our knowledge, no work has been done on coin die clustering.

\subsection{Point Cloud Registration} \label{sec:Point Cloud Registration}
Point Cloud registration\footnote{In this case, we mean rigid registration.} is still a hot topic in the literature and there are still a lot of challenges unsolved. Usually, registration methods are classified in two families: local registration and global registration. Local registration is an easier problem because we suppose that the 3D point cloud are almost aligned.
\paragraph{Local registration}
The most popular method in local registration is Iterative Closest Point \cite{besl1992icp,chen1992icp} and a lot of variants exist \cite{chen1992icp,Rusinkiewicz2001EfficientVO,pavlov_aa-icp:_2017,chetverikov_trimmed_2002}. For example, \cite{pommerleau2019robust} compares different cost functions, in order to make ICP robust to outliers. But in our case, we want to align patterns. Outliers are the border of the coins here so, the outliers are not random points in our case.
An other family of method uses probability density distribution (especially Gaussian mixture) to model the point cloud \cite{kanade_correlation-based_2004,jian2011gmm,myronenko_point-set_2010,horaud_rigid_2011,fleet_generative_2014,gao18filterreg,eckart_fast_2018}. If the point cloud is modeled as a Gaussian mixture, the transformation can be found using Expectation Maximization algorithm. The advantage of Gaussian mixture model methods is that it's more robust to noise and outliers. It is also possible to adapt the method to irregular point cloud, see \cite{Lawin_2018_CVPR} for more details. But the main drawback is the slowness of the method so it only works for small point cloud, even if recently some works \cite{gao18filterreg,eckart_fast_2018} have addressed this problem. Recently, some Deep Learning approaches like \cite{yaoki2019pointnetlk} or \cite{DBLP:journals/corr/abs-1905-04153} have been developed for point cloud registration (mainly registration of scene and 3D models). But the main drawback is that we need a lot of annotated data which is not our case.

\paragraph{Global registration} 
The previous methods only perform local registration ie the initialization needs to be closed to the right solution. 
Descriptors like FPFH \cite{hutchison_unique_2010}, SHOT \cite{hutchison_unique_2010}, Spin Image \cite{johnson_using_1999} can be used to compute matches between the two points cloud or more and then robust estimation (RANSAC or M-Estimator \cite{leibe_fast_2016}) can be performed to keep inliers and exclude outliers. GoICP \cite{yang_go-icp:_2016} performs branch and bound to compute the right transformation. It divides the space of solution into hypercube and then it computes a bound. Then with respect to the bound, it excludes part of the search space or it subdivides again the space. GOGMA \cite{campbell_gogma:_2016} performs branch and bound methods but models the point clouds using Gaussian Mixture models. \cite{ferrari_efficient_2018,Yang19tr-teaser} improves this method by searching only translations. \cite{Yang19tr-teaser} also use a truncated least square to be robust to outliers. It reduces the dimension of the solution search, so it is faster. Then, it is easier to find the right rotation. The main drawback with branch and bound methods is that a good bound is needed to exclude candidates. In the case of flat objects like coin, we can reduce the dimension of search space but it is more difficult to find the appropriate bound.
Discriminative optimization \cite{vongkulbhisal_discriminative_2017} and inverse discriminative optimization \cite{8578414} learn a sequence of update that lead to a stationary transformation instead of optimizing a cost function.

\section{METHOD}\label{sec:OUR METHOD}

To perform coin die recognition on a coin set, our method consists of 3 main steps:
\begin{enumerate}
	\item Pattern matching using registration 
	\item Automatic computing of probability that two coins come from the same die
	\item Coin clustering
\end{enumerate}
\begin{figure*}
	\includegraphics[width=\textwidth]{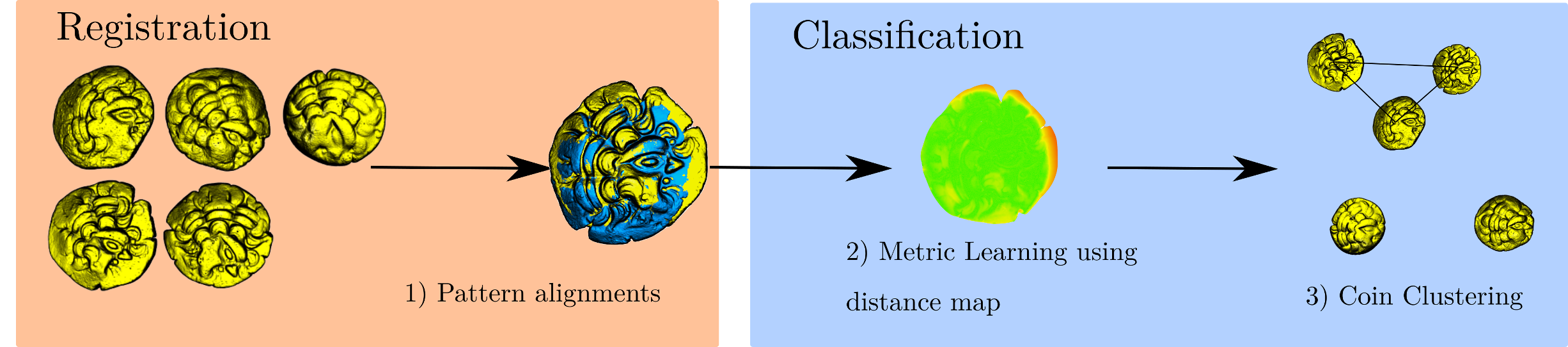}
	\caption{Pipeline of the method. First we perform registration to align patterns, then we compute a point to point distance map and we use machine learning to discriminate similar patterns from different patterns. Then we can group coins with similar patterns and coins with different patterns.}
	\label{fig:global_method}
\end{figure*}
The whole method can be summed up in Figure ~\ref{fig:global_method}.
% TODO figure of the method

\subsection{Registration algorithm}

\paragraph{Local Registration}
Let $X$ and $Y$, be point cloud of size $N$ and $M$, we want to align.
The goal is to find the translation and the rotation such that:
%TODO equation 
\begin{equation}
\label{equ:p2point}
R^*, t^* = \argmin_{R \in SO(3), t\in \mathbb{R}^3} \sum_{i=1}^{N}\sum_{i=1}^{M} \delta(x_i, y_j)||Rx_i+t -y_j||^2 
\end{equation}
where $\delta$ is a function which is equal to 1 if the points are matched and 0 if the points are not matched. In the case of Gaussian Mixture model, $\delta$ is a soft assignment function \cite{myronenko_point-set_2010}.
For the assignment between $X$ and $Y$, we usually use the nearest neighbor. Equation \ref{equ:p2point} can be rewritten as:
\begin{equation}
\label{equ:p2point2}
(R^{*}, t^{*}) = \argmin_{R \in SO(3), t \in \mathbb{R}^3} \sum_{i=1}^N||Rx_i+t-NN_{Y}(Rx_i+t)||^2
\end{equation}
and with:
\begin{equation}
\label{eq:loss_icp2}
NN_{Y}(x_i) = \argmin_{y \in Y} ||x_i -y||^2 
\end{equation}
Where $NN_Y$ is a function that compute the nearest neighbor on the point cloud $Y$.
The problem with the Equation \ref{equ:p2point2} is that we need the right matches to compute the correct transformation. The principle of ICP is the following:
\begin{itemize}
	\item We compute matches using the nearest neighbors.
	\item With the matches, we can compute the right transformation using the Kabsch algorithm.
\end{itemize}
Therefore, in ICP, at each iteration, we want to solve:
\begin{align}
y^{(n)}_i &= NN_{Y}(R^{(n)}x_i+t^{(n)}) \\
\label{equ:errOptim}
(R^{(n+1)}, t^{(n+1)}) &= \argmin_{R \in SO(3), t \in \mathbb{R}^3} \sum_{i=1}^N||Rx_i+t-y^{(n)}_i||^2
\end{align}
An interesting variant of ICP is point to plane ICP \cite{chen1992icp,Rusinkiewicz2001EfficientVO}. Instead of solving the Equation \ref{equ:errOptim}, we solve the following equation:
\begin{align}
y^{(n)}_i &= NN_{Y}(R^{(n)}x_i+t^{(n)})\\
n^{(n)}_{y_i} &\text{ is the normal of point $y^{(n)}_i$}\\
\label{equ:errL2plane}
(R^{(n+1)}, t^{(n+1)}) &= \argmin_{R \in SO(3), t \in \mathbb{R}^3} \sum_{i=1}^N((Rx_i+t-y^{(n)}_i).n^{(n)}_{y_i})^2 
\end{align}

Equation \ref{equ:errL2plane} has no analytic solution, but we can linearize the equation and use the Least Square Estimation to estimate the rotation and the translation. In that case, we need to suppose that the angle of rotation is small.

Due to the irregularity of the shape of each coin, we cannot apply ICP directly. Indeed, the border of coins will influence a lot the registration, whereas it doesn't belong to the patterns.
But we cannot use robust methods, because, here outliers are well organized shapes and not random points. 
Registration algorithms are generally sensible to the shape of the 3D Point Cloud. One way to deal with this problem is to simply remove the borders because in a coin, the pattern is usually at the center of the coin. To remove the border, we can compute the center of the coin and only keep the points that are within a ball of radius $r$. This simple trick allow ICP to be robust to borders.
\paragraph{Global registration}
As we've said in the previous sections, local registration works if the point cloud are partially aligned. It can be useful if we want to refine manual registration but if we want a fully automatic method, a global registration method is necessary.  The problem of ICP is that it finds a minimum of the equation \ref{eq:loss_icp2} but the minimum is local. 
If we want to find a global solution of the equation \ref{eq:loss_icp2}, we have several possibilities:
\begin{itemize}
	\item We can try random initializations and then for each initialization, perform an ICP. We'll call this method Random search ICP.
	\item We can try every initialization on a grid: we'll call it Grid Search ICP
	\item We can perform branch and bound optimization: it is GoICP \cite{yang_go-icp:_2016}.
\end{itemize}
We have tested the three methods. We will discuss the results in the experiments section, and we will see that Random Search ICP performs the best.
We have tested another algorithm of global registration based on descriptors. 
For the 3D descriptors, FPFH descriptors can be used for example. Then, we can compute the closest feature to obtain matches, and then filter the matches using the reciprocity test.
 With the matches, a robust method such as RANSAC  can be used to estimate the transformation and filter the correct matches.
For 2D descriptors, we can project the point cloud into plane, then SIFT descriptors \cite{lowe_distinctive_2004} can be used. 
As we will see in the experiment section, descriptors did not work on our dataset.

\subsection{Metric learning on patterns}
\begin{figure*}
	\includegraphics[width=\linewidth]{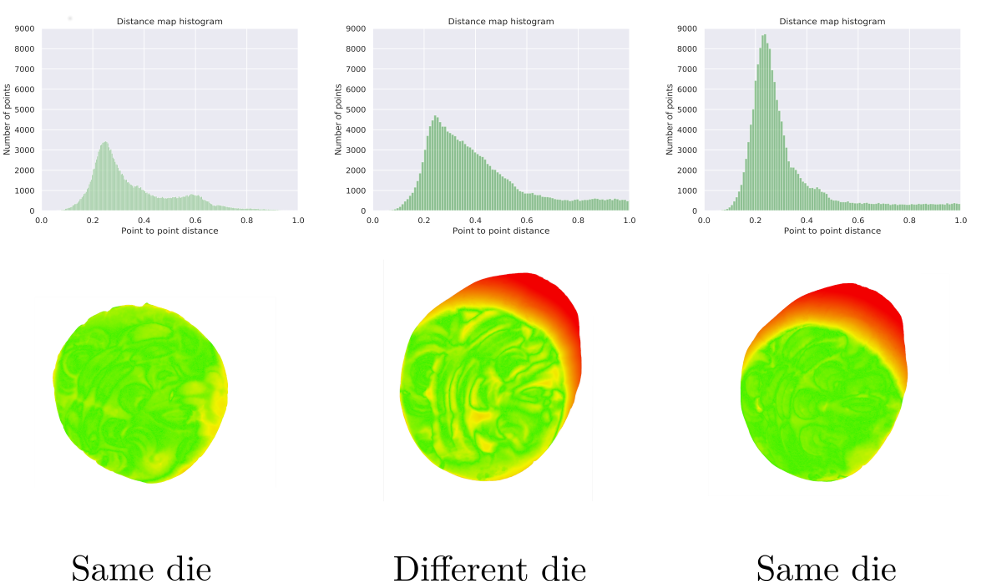}
	\caption{3 examples of distance map between two pair of coins on the bottom and their respective histogram on the top. }
	\label{fig:hist}
\end{figure*}

Now, we suppose that patterns of two coins are well aligned. First, we compute the point to point distance map $D$ between the two aligned point cloud $X$ and $Y$, using this formula:
\begin{equation}
D_i =  ||x_i - NN_Y(x_i)||^2
\end{equation}
Then, we compute the histogram of distance $h$ from $D$. $h$ is a distribution of the point to point distance between $X$ and $Y$. If the patterns are well aligned we should see a lot of points with small distances (see Figure \ref{fig:hist} for examples of point to point distance maps and their respective histograms). 
In order to decide whether the aligned patterns are the same or not, we propose to use machine learning procedure and just with the histogram of distance we can know if the pattern is the same or not. The machine learning algorithm will measure the probability that two patterns are similar. It can be seen as a metric learning procedure. 

\paragraph{Machine Learning algorithm}
We decide to use Logistic regression to solve our problem.
Let $h_1 \dots h_K$ be the histograms of a pair of coins (feature of dimension $d$) and let $l_1\dots l_K$ be their respective labels (1 if it is the same pattern, 0 if it is not). 
The objective is to find the parameters $\theta \in \mathbb{R}^d$ and $b\in \mathbb{R}$ such that:
\begin{equation}
\theta^*, b^* = \argmin_{\theta \in \mathbb{R}^d, b \in \mathbb{R}} \sum_{i=1}^{K}  \log(1 + \exp(-(\theta^Th_i + b)) - l_i(\theta^Th_i +b) + ||\theta||^2
\end{equation}
It can be minimized using gradient descent or Newton methods.
After finding the right parameters with the training set, we can evaluate quickly and automatically the probability that the patterns are similar. 
\subsection{Clustering of coins}
We have described a way to get the probability that two coins have the same die. With this data, we can cluster coins in different groups corresponding to the die they are coming from. 
Let $X_1,\dots X_n$ be the coins. We can now compute $p_{ij}$ the probability that $X_i$ and $X_j$ have the same pattern.
Let $P=(p_{ij})_{i=1,\dots, N, j=1\dots,N}$ the matrix of pair comparisons. $P$ can also be seen as a weighted graph where each node is a coin. To perform graph clustering, we can threshold graph weights to get a non weighted graph.
In other words, we compute :
\begin{equation}
A_{ij} = \mathbbm{1}_{p_{ij} \geq \alpha} = 
\begin{cases} 
1 & p_{ij}\geq \alpha \\
0 & p_{ij} < \alpha\\
\end{cases}
\end{equation}
 then we can compute the connected components of the graph $A$ to group coins with the same die pattern.

\section{EXPERIMENTS}\label{sec:EXPERIMENTS}
In this section, we will describe the experiments we've done to evaluate our method. The main questions is: How can we evaluate the whole pipeline ? To do so, we need to evaluate the components of the pipeline separately. We will first describe how the data have been acquired, then, we will explain how to evaluate each step of our method.

\subsection{Dataset}
The coins have been scanned with the AICON Smartscan (with surface measurement distance error around 50 $\mu$m). 3D Point clouds have around 300k points for one side of the coin. 881 coins have been scanned but only a small part of that dataset is annotated.
For 66 coins, we have all the pair transformations. We have therefore $\frac{N(N-1)}{2}$ comparisons ie 2145 comparisons for the metric learning. 
To evaluate the whole pipeline, we have 90 coins (we just need the labels to evaluate the coin die clustering).
Because annotating this treasury manually is complicated and tedious (we remind the readers that only an expert can annotate the data), we could not evaluate our method on the whole dataset.
 
%TODO figure of the label
\subsection{Implementation details} \label{sec:details of implementation}
We use Python and C++ to implement our method. We use Open 3D, a modern library in C++ to implement our own version of ICP. 
To perform logistic regression, we used scikit-learn with the default parameters. 
To plot the final graph, we use the Plotly library and Networkx library.
 
\subsection{How to evaluate registration?}
In order to justify the validity of our method, we manually registered point clouds and measured the angle error (in degree) and the translation error (in mm).
To measure errors, we use the following metrics:
% TODO FORMULA of the metric
\begin{align}
\label{equ:err_tr}
\delta t_{12} &= ||t_2 - t_1||_2 \\
\label{equ:err_rot}
\delta R_{12} &= ||\log_{SO3}(R_1R_2^T)||_2
\end{align}
%TODO typical error of a human table
By knowing the typical error of manual registration, we can correctly evaluate our method. 
If $\delta t_{12} < \delta t_{manual}$ and if $\delta R_{12} < \delta R_{manual}$, we can consider that the automatic registration succeed. For the rest of the experiments, we'll use $\delta R_{manual}=5$ degrees and $\delta t_{manual}=0.5mm$ which is a good order of magnitude.

\begin{figure*}[ht]
	\centering
	\begin{subfigure}{.45\textwidth}
		\centering
		\includegraphics[width=0.8\linewidth]{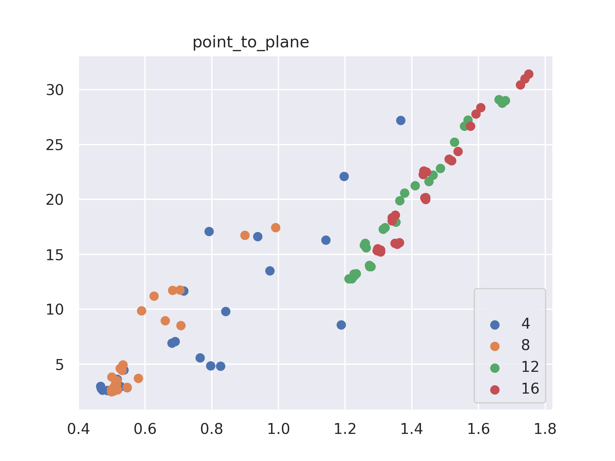}	
		\caption{Point to Plane ICP}
	\end{subfigure}
	~	
	\begin{subfigure}{.45\textwidth}
		\centering
		\includegraphics[width=0.8\linewidth]{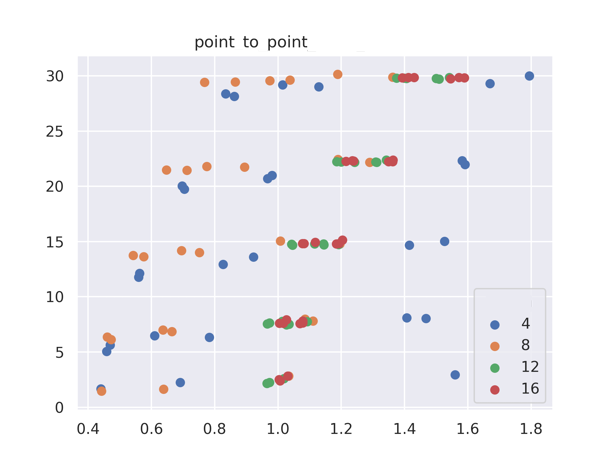}
		\caption{Point to Point ICP}
	\end{subfigure}
	
	\caption{Translation error (absciss axis) and rotation error (ordinate axis) with different sizes of border exclusion (the radius is in cm). Each point represents the mean of error on pairs of coin with 100 random initializations. We can remark that Point to Point ICP doesn't converge. Point to Plane ICP converges only if we exclude the border but if we exclude too much, it doesn't work. The best radius in our experiments is 8 mm.}
	\label{fig:preuve_icp}
\end{figure*}

\paragraph{Evaluation of local registration}
To evaluate local registration, we've used 100 pairs of coins with their transformations computed manually.
\begin{figure}[ht!]
	\centering
	\includegraphics[width=0.54\textwidth]{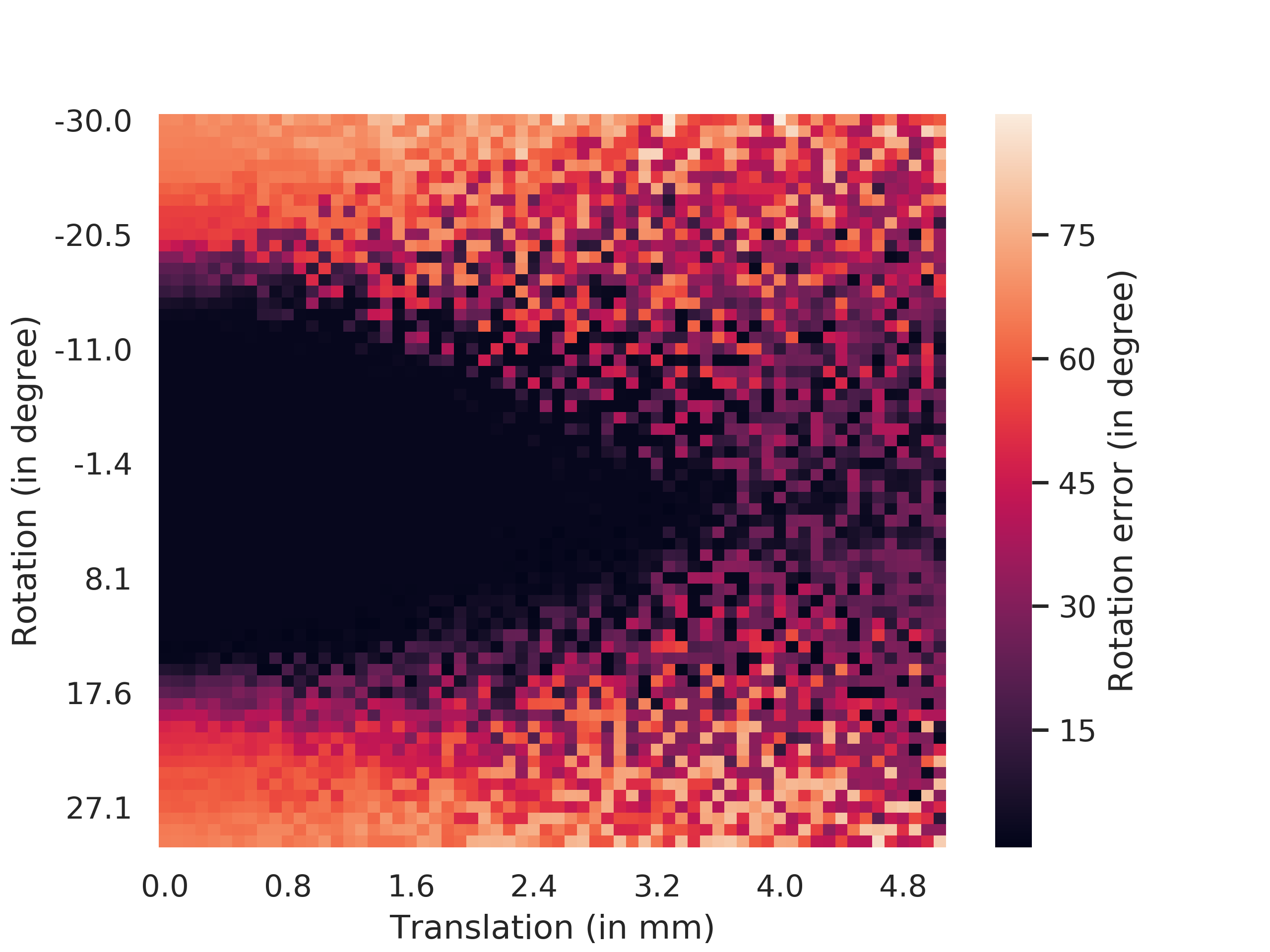}
	\caption{Convergence basin of ICP with respect to the initialization for pair of coin coming from the same die. In absciss axis, the norm of the translation varies from 0 to 5 mm. In ordinate axis, the rotation varies from -30 to 30 degrees. The dark part in this map represents the convergence basin of ICP.}
	\label{fig:convicp}
\end{figure}
For the local registration problem, we've evaluated different variants of the ICP to see which parameters are important:
\begin{enumerate}
	\item The influence of border
	\item The influence of the cost function (Point to Point or Point to Plane)
\end{enumerate}
We can see in Figure \ref{fig:preuve_icp} the influence of border exclusion and that point to plane ICP is better than point to point ICP. Indeed, according to \cite{Rusinkiewicz2001EfficientVO}, point to plane ICP work very well on flat structures. By removing the borders of the coins, we have lower errors. But for some coins, we can see that error is still big even with point to plane and by removing borders. Indeed, if the patterns are not similar (coming from different dies), the registration will fail because we're trying to registrate different patterns. But if the patterns are coming from the same die, by using the criterion defined above, \textbf{98\%} of coins are correctly registered. An ICP can be done in approximately 300 ms. For all other experiments, we keep the best radius for border exclusion as 8 mm.

\paragraph{Influence of the initialization}
To evaluate the influence of the initialization, we tried to compute the convergence basin of ICP. To what extent does ICP converge to the right solution? We can see in Figure \ref{fig:convicp}, a visualization of the convergence basin by varying two parameters: for the translation, we've taken a random direction and the norm of this translation has been varied. Also, the rotation around the $z$ axis has been varied. We can conclude that if the distance between the initialization and the global minimum is less that 10 degrees in rotation and 3 mm in translation, we are nearly sure that we will converge with ICP. This information is very useful for the global registration.  
\paragraph{Comparison between different global methods}
We've experimented different algorithms for the global registration.

\begin{table*}[ht]
	\centering
	\begin{tabular}{|l|l|l|l|}
		%\toprule
		\hline
		{} &  Error $<$ 1 deg &  Error $<$ 5 deg &  Error $<$  0.5 mm \\ \hline
		%\midrule
		2D descriptors 					& $<$1\%       &       $<$1\%& $<$1\% \\ \hline 
		3D descriptors 					& $<$1\%       &       $<$1\%& $<$1\% \\ \hline 
		
		grid search        &       18\% &       84\% &            75\% \\ \hline
		goICP        &       18\% &       84\% &            75\% \\ \hline
		random search      &       \textbf{22\%} &       \textbf{94\%} &            \textbf{86\%} \\ \hline
		%\bottomrule
	\end{tabular}
	\caption{Success rate in \% of global registration. We measure the error using Equation \ref{equ:err_tr} and \ref{equ:err_rot}. For grid search and random search, we perform the same number of trial (hundred of trials). GoICP has the same results as grid search if we limit the number of iterations for GoICP.}
	\label{table:success}
\end{table*}
Because, we work on flat point clouds, we can reduce the dimension of space search (instead of 6 degrees of freedom, we have 3 degrees).
In Table \ref{table:success}, we can see which global algorithms converge to the right solution. The descriptors approach doesn't work here. One possible reason is that the coin contains repetitive patterns. The descriptors are too local therefore, it is hard to take into account the global structure of the pattern.
\begin{figure*}[b]
	\centering
	\includegraphics[width=0.7\linewidth]{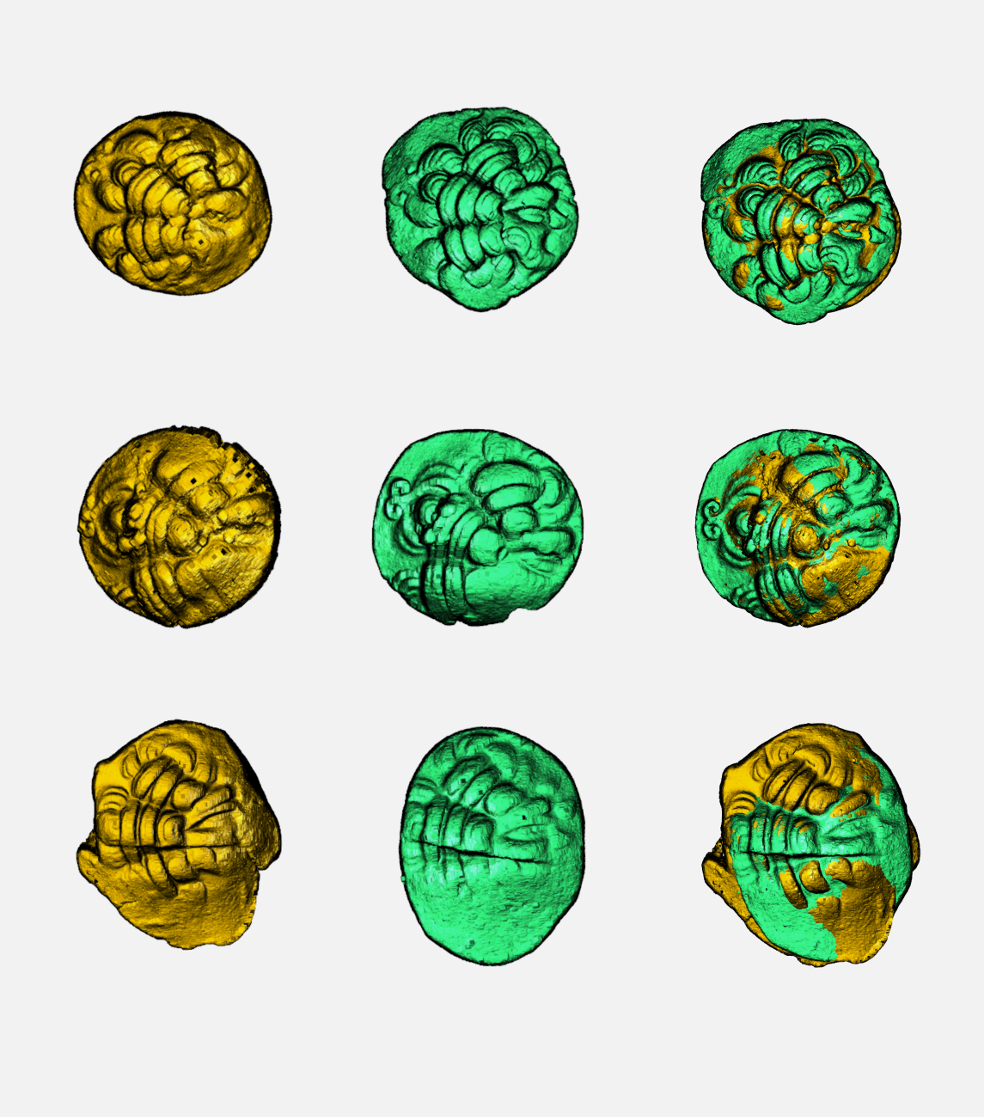}
	\caption{Qualitative results of global registration of pair of coins using Random Search ICP.}
	\label{fig:quali}
\end{figure*}
 Even when there are some missing parts or when the coin is partially damaged, the registration still succeed in aligning patterns. GoICP performs also well but it is too slow, we've realized that GoICP was equivalent to Grid Search because, the bound was not accurate enough to exclude some candidates. Random search here has the best results. In Figure \ref{fig:quali}, we can see the registration of pair of coins using Random Search ICP.

\subsection{Metric learning evaluation}
We have seen in the previous section that the decision whether two coins have the same die is a problem of binary classification. To correctly evaluate this part, we have aligned by hand the patterns, then we perform an ICP to refine and then, we can test if two patterns are similar or not. 
As it said in previous section, the dataset is composed of 2145 comparisons so 2145 histograms. In our dataset, we have 1436 negative labels and  709 positive labels. We divide our dataset randomly in a training set and a test set: half of our dataset is used for training, the other half is used for testing.
We can simply compute the accuracy on the test set to see if our method is correct or not. 
When the patterns are correctly registered using ICP, the accuracy is \textbf{98\%} on the test set. But if we registrate the patterns by hand the accuracy drop to \textbf{72\%} on the test set. It shows that ICP allows a better repeatability in decision. 
We can see that the method is very effective when the patterns are well aligned. When patterns are the same, the histograms of distances are good features to see if the patterns are similar or not. But it will fail when the coins are not well aligned.
We have also shown that the metric learning part doesn't need a huge dataset to learn an appropriate metric.
\begin{figure*}
	\includegraphics[width=\linewidth, height=0.3\textheight]{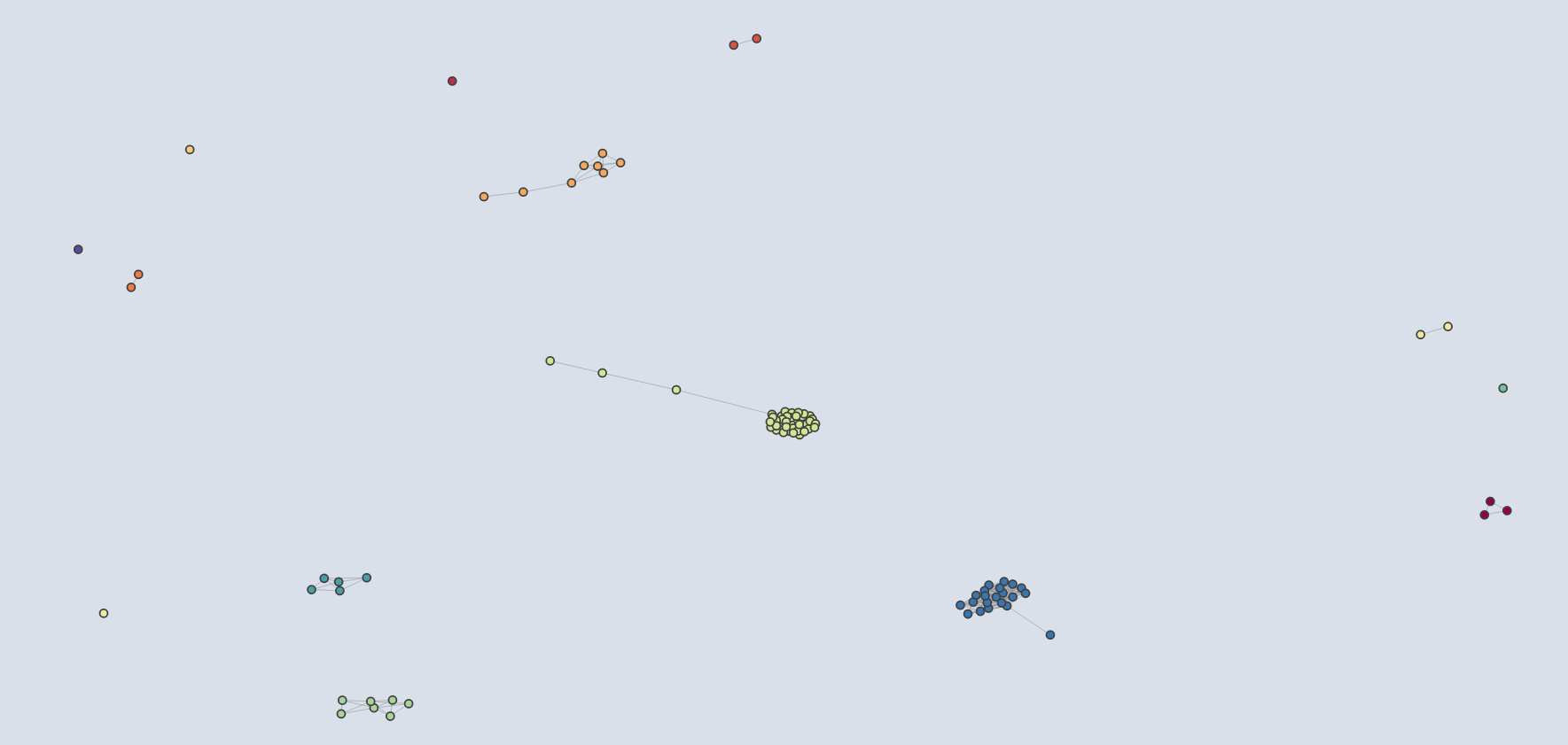}
	\caption{Clustering of 90 coins. Each node represents a coin and coins of the same cluster are linked together and have the same color.}
	\label{fig:graph}
\end{figure*}
 
 \subsection{Clustering evaluation}
 We did the evaluation on 90 labeled coins.
 This part is the most difficult part to evaluate because, the treasury is not fully labeled yet and also, the archaeologists could not have done all the comparisons between coins. We can see in Figure \ref{fig:graph} a representation of the graph. Each vertex is a coin and the connected components are a die. Normally, every cluster should be a complete graph (ie all coins in a cluster connected two by two), but the registration can sometimes fail. Also, if the coin is damaged, we can see that the connected component graph will not be complete. Indeed, if a coin is damaged or fractured, the algorithm can fail recognizing the die. Quantitatively, there are several methods to evaluate our clustering method,we decide to use the adjusted rand index (ARI) \cite{Hubert1985}.
 It is defined as :
 \begin{equation}
 ARI = \frac{RI - E[RI]}{\max(RI) - E[RI]}
 \end{equation}
 $RI$ is called the rand index. $ARI$ is more relevant because, for random labels, the value is closed to 0. The best value we can have is 1.
 In our case, the ARI has a value of \textbf{0.86}. Most of the time, the method is correct but there are still some mistakes.  
 \paragraph{Where do the mistakes come from ?}
 As we have said earlier, some coins are damaged or even fractured, it is therefore more difficult to registrate the patterns. Also, the metric learning method can indicate that two coins with the same die are different, because one coin is too damaged or has a fracture. These particular cases explain why there are still some mistakes and each sub-graph is not always complete in Figure \ref{fig:graph}. Even if there are still some errors, the clusters are still meaningful and can be useful for experts.

\section{CONCLUSION AND DISCUSSIONS}
This work introduces the challenging problem of coin die clustering and a method to solve it on real data. To our knowledge, no  automatic method exists in the literature to cluster patterns on 3D point cloud that only experts can identify. To do so, we can perform coin registration no matter the shape of coins. Then, with small annotated dataset, we can learn an accurate metric which gives a probability that two coins have the same die. Finally, we can obtain clusters of coins and, they are relevant for archaeologists. This work will allow them to analyze automatically treasuries and to know how many dies have been used at that time. In future work, we are interested in generalizing pattern recognition on every kind of shapes and not just coins.
{
	\begin{spacing}{1.17}
		\normalsize
		\bibliography{ISPRSguidelines_authors} % Include your own bibliography (*.bib), style is given in isprs.cls
	\end{spacing}
}

\vspace{1cm}

\end{document}